\documentclass{article}

\usepackage{PRIMEarxiv}

\usepackage[utf8]{inputenc} 
\usepackage[T1]{fontenc}    
\usepackage{hyperref}       
\usepackage{url}            
\usepackage{booktabs}       
\usepackage{amsfonts}       
\usepackage{nicefrac}       
\usepackage{microtype}      
\usepackage{lipsum}
\usepackage{fancyhdr}       
\usepackage{graphicx}       
\graphicspath{{media/}}     
\usepackage[ruled,linesnumbered]{algorithm2e}
\usepackage{lineno,hyperref}
\usepackage{subfig}
\usepackage{amsmath}
\usepackage{amssymb}
\usepackage{amsfonts}
\usepackage[table]{xcolor}
\usepackage{graphicx}
\usepackage[english]{babel}
\usepackage[utf8]{inputenc}
\usepackage[noend]{algpseudocode}
\usepackage{color,soul}
\modulolinenumbers[1]
\usepackage{lineno}
\usepackage[symbol]{footmisc}
\title{UnfairGAN: An Enhanced Generative Adversarial Network for Raindrop Removal from A Single Image
}

\author{
  Duc Manh Nguyen \\
  Vietnam National Space Center  \\
  Vietnam Academy of Science and Technology \\
  Hanoi, Vietnam\\
  \texttt{manh.nguyenduc.uet@gmail.com} \\
   \And
  Sang-Woong Lee \\
  Pattern Recognition and Machine Learning Lab \\
  Gachon University \\
  Seoul, South Korea\\
  \texttt{slee@gachon.ac.kr} \\
}

\begin{document}
\maketitle

\begin{abstract}
Image deraining is a new challenging problem in real-world applications, such as autonomous vehicles. In a bad weather condition of heavy rainfall, raindrops, mainly hitting glasses or windshields, can significantly reduce observation ability. Moreover, raindrops spreading over the glass can yield refraction's physical effect, which seriously impedes the sightline or undermine machine learning systems. In this paper, we propose an enhanced generative adversarial network to deal with the challenging problems of raindrops. UnfairGAN is an enhanced generative adversarial network that can utilize prior high-level information, such as edges and rain estimation, to boost deraining performance. To demonstrate UnfairGAN, we introduce a large dataset for training deep learning models of rain removal. The experimental results show that our proposed method is superior to other state-of-the-art approaches of deraining raindrops regarding quantitative metrics and visual quality. 
\end{abstract}


\section{Introduction}

In a scene, rain at different levels creates a compact set of visual effects, leading to vision systems failures such as object detection and recognition. Although many approaches \cite{KGarg2004} \cite{KGarg2007} \cite{SHasirlioglu2018} \cite{HKurihata2005} \cite{HKurihata2007} \cite{MRoser2009} \cite{MRoser2010} \cite{AYamashita2003} \cite{AYamashita2005} \cite{GVolk2019} \cite{SYou2013} have focused on eliminating the adverse impacts of raindrops, deraining images or videos is still challenging for three reasons. First, restoring information degraded by raindrops is an ill-posed problem because raindrops vary in a wide variety of sizes, shapes, and their appearance highly depends on the surrounding environment. It is hard to find an appropriate physical model of raindrops on the glass of the window. Second, heavy raindrops, dropping from the sky at high speed, hit and spread the surface of windows glasses or car windshields, creating unpredictable water flows and causing fuzzy backgrounds. Furthermore, the focus of a vision-based system is mainly aimed at the background and objects appearing in the scene, making raindrops blurred \cite{RQian2018}. These blurred raindrops can either fully occlude the scene behind or partly remove some details from the occluded background. Finally, thick raindrops spreading over the glass pane or windshield cause the physical effect of refraction \cite{SYou2013} \cite{JCHalimeh2016}, making the vision-based system unable to accurately estimate shapes and real positions of objects. 
 
 \begin{figure*}
 \centering
 \includegraphics[width=1\linewidth]{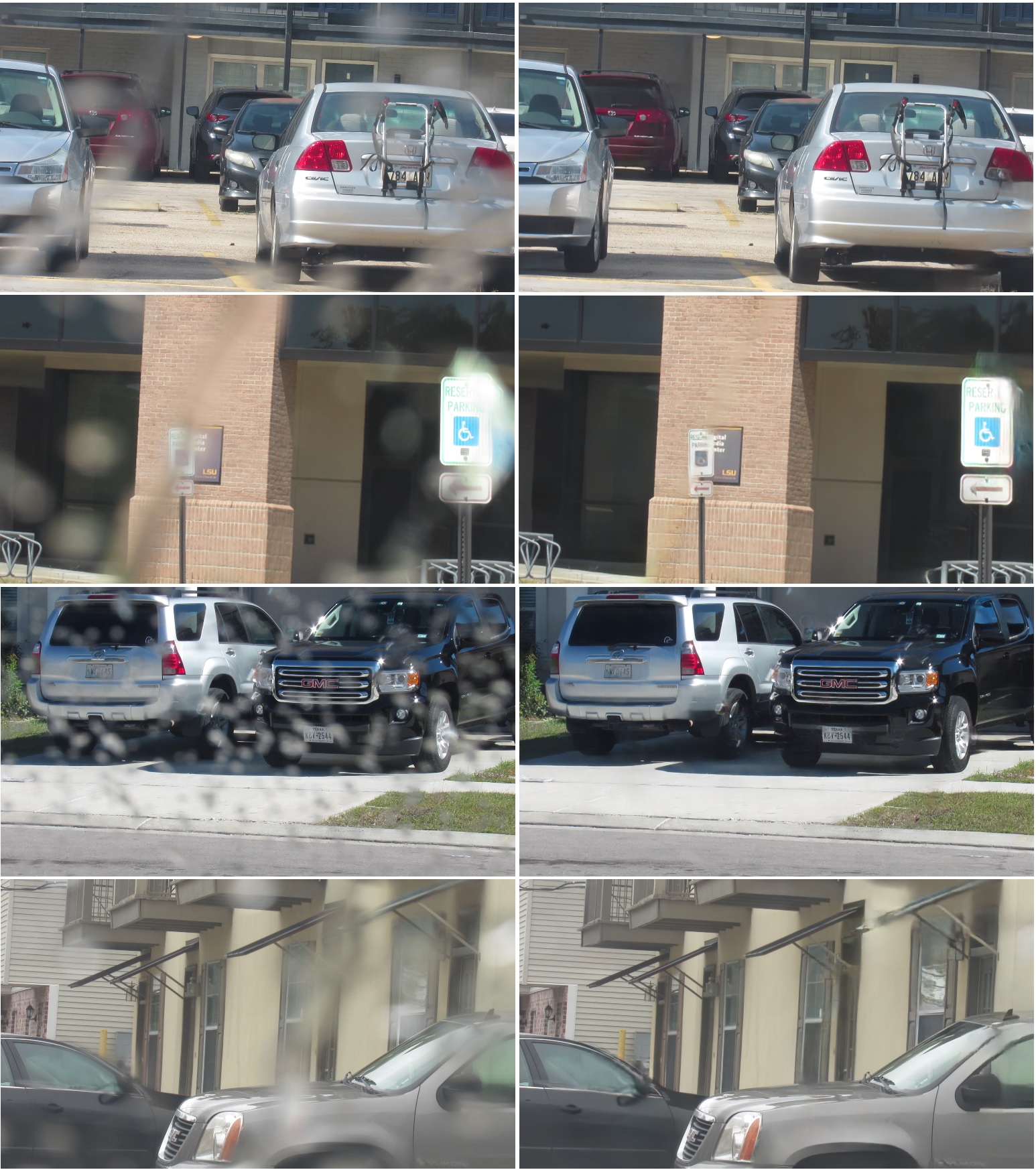} 
 \caption{Demonstration of our proposed method. Column 1: Original images with real-world raindrops and rain flows. Column 2: Our results.}
 \label{fig_derain_0}
\end{figure*}

Inspired by state-of-the-art image deraining methods \cite{SYou2013} \cite{RQian2018}, we propose an enhanced generative adversarial network (UnfairGAN) to address raindrops' physical problems. First, UnfairGAN can eliminate different kinds of raindrops while recovering texture details and object edges. Recently, deep convolutional neural networks (DCNN) \cite{YQuan2019} have shown their superior performance in removing rain. However, they tend to over-smooth raindrops and erase details on different local areas of the image. UnfairGAN is highly effective in recovering the lost information and edges caused by raindrops thanks to a proposed feature attention module and prior high-level information maps. Two prior high-level information maps, including an edge attention map and a rain estimation map, are extracted from two corresponding estimation networks. To design a useful feature attention module, we propose using a new dilated activation function (DAF) to improve the attention mechanism's effectiveness. Second, UnfairGAN is superior to other competing methods in addressing heavy raindrops, causing complicated and broad shapes and refraction's physical effect. UnfairGAN can produce reconstructed images closing to natural photos even though the refraction effect severely demolish object details. Finally, to improve the effectiveness of the generator of UnfairGAN, we propose a novel dilated residual dense UNets (DRD-UNet) combining the advantages of dilated residual networks \cite{FYu2017}, and UNets \cite{Ronneberger2015}. DRD-UNet can effectively extract hierarchical features for image restoration at different scales. In short, our main contributions are:

\begin{itemize}

\item We propose a new conditional GAN (UnfairGAN) that can effectively preserve the essential details caused by heavy raindrops and eliminate artifacts caused by instability of training the discriminator.
 
\item We introduce a novel advanced attention module (AAM) that can effectively extract different kinds of prior information to boost reconstruction performance. 
 
\item We present a new advanced activation function that can boost the learning effectiveness of image classification and reconstruction. 

\item We present DRD-UNet that can probe hierarchical features for image restoration effectively. DRD-UNet is an end-to-end cascade network that can boost the performance of reconstructing realistic rain images. 

\item We introduce a new large image dataset (DeepRaindrops) for training deep learning networks of removing raindrops, as seen in Figure \ref{fig_derain_0}. 

\end{itemize}

\section{Related Work}
 
\begin{figure*}
 \centering
  \includegraphics[width=1\linewidth]{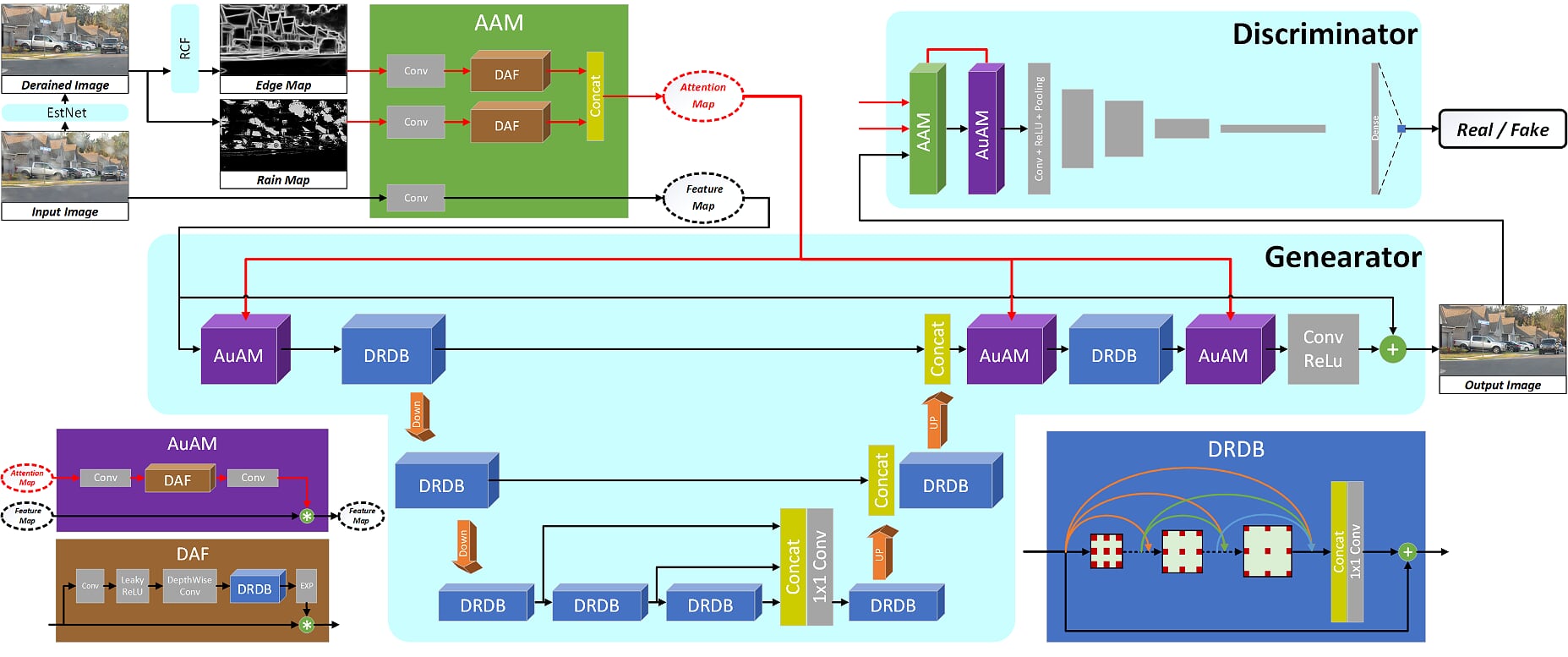} 
 \caption{Block diagram of the architecture of UnfairGAN.}
 \label{fig_resnet}
\end{figure*}

Recently, there has been more and more demand for eliminating raindrops caused by bad weather \cite{RTTan2018} \cite{SZang2019} \cite{FBernardin2013}. You et al. \cite{SYou2013} \cite{SYou2016} analyzed the derivative properties of raindrops and presented two different physical models of clear adherent raindrops and blurred raindrops. Interestingly, Quan et al. \cite{YQuan2019} proposed to extract edges based on the mathematical model of raindrops' shape. This method obtained remarkable performance of removing raindrops on the Raindrops dataset \cite{RQian2018}. However, this model assumes that the raindrop is relatively round, only valid with tiny raindrops. Raindrops can hit and spread mainly on the glass pane's surface, causing unpredictable water flows and complex raindrops with various shapes and sizes. This assumption is not right in real-world environments. Since the Raindrops dataset \cite{RQian2018} has not enough training images to train a deep learning model for deraining tasks effectively, Hao et al. \cite{ZHao2019} introduced a new dataset of the photo-realistic rendering of raindrops, including 30,000 training pairs. The rendering is derived from physics-based water dynamics, water geometric, and photometry. However, this physical model of raindrops is not matched the real-world raindrops model. Thus, the proposed 3RN method in this paper is not an effective deraining method for real-world applications. Finally, Li et al. \cite{RLi2020} proposed a reconstruction method using a neural architecture search to identify and handle multiple weather types, including rain, fog, snow, and raindrops. Since this method has to deal with a wide variety of degradations, its reconstruction performance is limited.

\section{Methodology}

In practical applications, successful algorithms should attain improved visual quality and high accuracy to generate deraining images that look like real images. Thus, we develop UnfairGAN to resolve the challenging problems of raindrops. Specifically, in UnfairGAN, the generative network $\mathbf{G}$ competes with the discriminator $\mathbf{D}$ to generate derained images, which can preserve textures and edges while also having a high PSNR value. A GAN-based method's primary goal is to obtain a Nash equilibrium to a min-max game between two players, including a generator $\mathbf{G}$ and a discriminator $\mathbf{D}$. Each player competes with its opposer to minimize its cost function. However, finding Nash equilibria is a challenging problem for generative adversarial networks. It is because it requires finding a training strategy that balances the generator's capacities and the discriminator. Unfortunately, in many image reconstruction cases, the discriminator often overpowers the generator, causing the training process's instability.
Consequently, the discriminator quickly converges to zeros even though the generator does not find the point that minimizes its cost function. The generator does not find any new update from the discriminator to learn. One more important thing is that the generator can even fool the discriminator by only generating bad fake samples. In turn, the discriminator becomes unstable and loses its ability to evaluate samples from the generator. For these reasons, we propose UnfairGAN that can boost both the generator and the discriminator's performances by making the zero-sum game between them fair enough.

\subsection{Generative Adversarial Networks}

To improve the discriminator's stability, UnfairGAN aims to make the training process of the discriminator more challenging. Since the discriminator firmly surpasses the generator in most cases, we force the discriminator to play the minimax game with a more strict rule (an unfair rule) than the generator. Thus, the minimax game between them becomes more competitive and challenging. In the new rule, the discriminator is forced to distinguish between a real image and a fake image with increasingly higher quality than the image generated directly from the generator. As a result, the discriminator is trained harder and harder over time instead of converging quickly with bad fake images. Given $\tilde{x}=(x_{r},x_{f})$, which is a pair of real and fake images, we generate an advanced fake image $x_{f}^{*}$ to update the discriminator $\mathbf{D}$ by the following formula. 

\begin{equation}
x_{f}^{*}=x_{f}+\mu\odot\eta \odot(x_{r}-x_{f})
\end{equation}
where $\odot$ is the pixel-wise multiplication operator, $\mu\sim U(0,1)$ is the learning factor, and $\eta \sim [0,1]$ is the random factor having the standard uniform distribution with minimum 0 and maximum 1. Recent advanced GAN-based methods \cite{SJenni2015} \cite{TSalimans2016} \cite{Vo2021a} \cite{Vo2021b} have a similar strategy to increase the accuracy of the discriminator. In these methods, Gaussian noise or instance noise is added to the discriminator's input to stabilize networks' data distributions. Since noise is highly sensitive to image denoising networks, we employ random $x_{f}^{*}$ to force the discriminator to train data harder and more precise. Applying an unfair rule to the discriminator is beneficial for the overall stability of both competing networks.

Jolicoeur-Martineau \cite{Martineau2018} demonstrated that GAN-based methods using a relativistic discriminator are more stable than standard ones. Thus, we use the formula $\mathbf{D}_{F}(\tilde{x})=sigmoid(\mathbf{C}(x_{r})-\mathbf{C}(x_{f}))$ to make the discriminator $\mathbf{D}$ relativistic and more stable, which $\mathbf{C}(x)$ is the output non-transformed layer of $\mathbf{D}$. In fact, $\mathbf{D}(\tilde{x})$ is the probability that the fake image $x_{f}$ is more realistic than its corresponding ground-truth image. Particularly, the discriminator loss function of UnfairGAN can be computed by the following formulas:

\begin{equation}
\begin{aligned}
\mathbf{L}_{D}^{adv}(x_{f}^{*}, x_{r})=&\mathbf{E}_{x_{r}\sim \mathbb{P}}\begin{bmatrix}
(\mathbf{C}(x_{r})-\mathbf{E}_{x_{f}^{*}\sim \mathbb{Q}}\mathbf{C}(x_{f}^{*})-1)^{2}
\end{bmatrix} \\
+&\mathbf{E}_{x_{f}^{*}\sim \mathbb{Q}}\begin{bmatrix}
(\mathbf{C}(x_{f}^{*})-\mathbf{E}_{x_{r}\sim \mathbb{P}}\mathbf{C}(x_{r})+1)^{2}
\end{bmatrix}
\end{aligned}
\end{equation}
where $\mathbb{P}$ is the distribution of real clean images, $\mathbb{Q}$ is the distribution of fake derained images. The adversarial loss function for the generator is computed by:

\begin{equation}
\begin{aligned}
\mathbf{L}_{G}^{adv}(x_{f}, x_{r})=&\mathbf{E}_{x_{f}\sim \mathbb{P}}\begin{bmatrix}
(\mathbf{C}(x_{f})-\mathbf{E}_{x_{r}\sim \mathbb{P}}\mathbf{C}(x_{r})-1)^{2}
\end{bmatrix} \\
+&\mathbf{E}_{x_{r}\sim \mathbb{P}}\begin{bmatrix}
(\mathbf{C}(x_{r})-\mathbf{E}_{x_{f}\sim \mathbb{Q}}\mathbf{C}(x_{f})+1)^{2}
\end{bmatrix}
\end{aligned}
\end{equation}
Based on this function, the generator is trained smoothly and generates the output data that is more realistic and closer to real data. In the next sections, we continue to introduce conditional UnfairGAN that utilizes prior information to improve further the quality of derained images.

\subsection{Generator Network}

Given a rain image $\mathbf{O}\in \mathbb{R}^{C\times W\times H}$, we train a generator to estimate the background image $\mathbf{\hat{I}}_{dr}^{G}\in \mathbb{R}^{C\times W\times H}$ so this it is as similar to its ground truth image $\mathbf{I}_{gt}\in \mathbb{R}^{C\times W\times H}$ as possible. To deal with the challenging problems of raindrops, we aim to explore reliable prior information which is able to help generative networks restore natural objects and detail, such as buildings, roads, trees. In this paper, two kinds of prior information are used: the attention rain map, $\mathbf{M}_{R}$, and the edge map, $\mathbf{M}_{E}$. Thus, the generator $\mathbf{G}$ is trained to directly map the rain image $\mathbf{O}$ to the estimated background image $\mathbf{\hat{I}}_{dr}^{G}$ as follows:

\begin{equation}
\mathbf{\hat{I}}_{dr}^{G}=\mathbf{G}(\mathbf{O}|\mathbf{M}_{R}, \mathbf{M}_{E})
\end{equation}
$\mathbf{M}_{R}$ can be directly generated from the output of a rain estimation network (EstNet). Indeed, $\mathbf{M}_{R}$ helps $\mathbf{G}$ focus on local rain regions and learn to eliminate complex raindrops better. Similarly, $\mathbf{M}_{E}$ is an output of the RCF network \cite{YLiu2017}. Raindrops can occlude and deform several local areas, causing details and edges in these areas to be blurred or entirely lost. Moreover, although conventional deep convolution neural networks can achieve high PSNR values, they often generate the over-smoothing effect, leading to the loss of high-frequency details and low-contrast features. In this paper, $\mathbf{M}_{E}$ becomes vital prior information to enhance the quality of the derained image. By paying attention to edge areas indicated by $\mathbf{M}_{E}$, $\mathbf{G}$ can preserve more textures and edges while achieving high PSNRs and SSIMs. To effectively update prior information to the generator, we develop an advanced attention module (AAM), which is explained in detail in the next section. Our network is illustrated in Figure \ref{fig_resnet}.

\subsection{Advanced Attetnion Module}

AAM is designed to map prior knowledge into complementary feature maps adding to original feature maps. This mapping can be presented as follows:

\begin{equation}
\mathbf{AAM}: (\mathbf{M}_{R}, \mathbf{M}_{E})\rightarrow \mathbf{M}_{F}^{*}
\end{equation}
where $\mathbf{M}_{F}^{*} \in \mathbb{R}^{C_{F}\times H\times W}$ is a complementary feature map corresponding to the attention maps $\mathbf{M}_{R} \in \mathbb{R}^{C_{R}\times H\times W}$ and $\mathbf{M}_{E} \in \mathbb{R}^{C_{E}\times H\times W}$. The architecture of the AAM is shown in Figure \ref{fig_resnet}. $\mathbf{M}_{F}^{*}$ can be generated by using the following formula:

\begin{equation}
\mathbf{M}_{F}^{*} = \mathbf{\Psi} (\mathbf{\Phi}(\mathbf{M}_{R} \circledast \mathbf{K}_{R}), \mathbf{\Phi}(\mathbf{M}_{E} \circledast \mathbf{K}_{E}))
\end{equation}
where $\mathbf{\Psi}$ is the concatenation operator, $\circledast$ is the convolution operator. Additionally, $\mathbf{K}_{R}$ and $\mathbf{K}_{E}$ are the corresponding sets of convolution kernels. Moreover, $\mathbf{\Phi}$ is the learnable dilated activation function (DAF) that is helpful in encoding complex structures of prior information. Finally, we use $N$ auxiliary attention modules (AuAM) to keep mapping $\mathbf{M}_{F}^{*}$ into auxiliary feature maps, $\mathbf{T}_{k}^{*}$, corresponding to $N$ hidden feature maps, $\mathbf{M}_{k}$, in $\mathbf{G}$, with $k=1,2,..,N$. $\mathbf{T}_{k}$ is the map that we aim to update prior information. The mapping between $\mathbf{M}_{F}^{*}$, $\mathbf{T}_{k}^{*}$ and $\mathbf{T}_{k}$ can be presented as follows:

\begin{equation}
\mathbf{AuAM}: (\mathbf{M}_{F}^{*}, \mathbf{T}_{k}) \rightarrow \mathbf{T}_{k}^{*}, \, \, \, \, \, k=1,2,..,N
\end{equation}
Each $\mathbf{T}_{k}^{*}$ is computed by the following formula:

\begin{equation}
\mathbf{T}_{k}^{*}=\mathbf{T}_{k} \odot (\mathbf{\Phi} (\mathbf{M}_{F}^{*} \circledast \mathbf{K}_{k}^{f})\circledast \mathbf{K}_{k}^{s})
\end{equation}
where $\odot$ is the element-wise multiplication operator. $\mathbf{K}_{k}^{f}$ and $\mathbf{K}_{k}^{s}$ are the sets of convolution kernels corresponding to $\mathbf{M}_{k}$.

\subsection{Dilated Activation Function}

DAF is an advanced activation function that can boost the effectiveness of classifying prior information, including edges and rain estimation. DAF aims to extract edges, rain estimation, and their derivatives. Unlike other element-wise activation functions such as ReLU, DAF can search gradient values on neighborhoods of each pixel in the image by using a dilated residual dense block inside (DRDB). DRDB allows the advanced attention module to observe multiple receptive fields and maximize information flow to pass through the activation function layer. By integrating DRDB into an activation layer, AAM can effectively probe hierarchical features and is far more substantial than traditional networks that separately include convolutional layers and element-wise activation functions. Figure \ref{fig_resnet} illustrates the architecture of DAF with an input $\mathbf{v}$ and an output $\mathbf{u}$. The output $\mathbf{u}$ can be calculated by:

\begin{equation}
\mathbf{u} = \mathbf{\Phi}(\mathbf{v})
\end{equation}
The activation function layer, $\mathbf{\Phi}$, is trained together with $\mathbf{G}$ to generate a weight map $\mathbf{w}$. Eq. (9) can be rewritten as 

\begin{equation}
\mathbf{u} = \mathbf{v} \odot \mathbf{w}
\end{equation}
Then the weight map $\mathbf{w}$ is computed by 

\begin{equation}
\mathbf{w} = \exp (-(\mathbf{f}_{1}(\mathbf{f}_{2} (\mathbf{f}_{3} (\mathbf{v}))))^{2})
\end{equation}
where $\mathbf{f}_{1}$, $\mathbf{f}_{2}$ and $\mathbf{f}_{3}$ are the depth-wise convolution layer, DRDB and the LeakyReLU function, respectively.

\subsection{Training of The Generator}

To increase PSNR and adaptively preserve essential details in the original image, the loss function of $\mathbf{G}$ is presented by:

\begin{equation}
\mathbf{L}_{G}=\mathbf{\omega} _{1}\mathbf{L}^{l_{1}} + \mathbf{\omega} _{2}\mathbf{L}^{MS\textsc{-}SSIM} + \mathbf{\omega} _{3}\mathbf{L}_{G^{adv}} + \mathbf{\omega} _{4}\mathbf{L}_{per}
\end{equation}
which $\mathbf{\omega} _{1}$, $\mathbf{\omega} _{2}$, $\mathbf{\omega} _{3}$, $\mathbf{\omega} _{4}$ are the regularization parameters. In this paper, we set $\mathbf{\omega} _{1}=0.16$, $\mathbf{\omega} _{2}=0.84$, $\mathbf{\omega} _{3}=0.001$, and $\mathbf{\omega} _{4}=0.01$. Additionally, $\mathbf{L}^{l_{1}}$ is the mean absolute error (MAE) loss, $\mathbf{L}^{MS\textsc{-}SSIM}$ is the loss of multi-scale structural similarity index \cite{HZhao2017}, $\mathbf{L}^{per}$ is the perceptual loss function generated by the DenseNet network \cite{GHuang2017}, and $\mathbf{L}_{G}^{adv}$ is the adversarial loss, as presented in Eq. (3). $\mathbf{L}^{l_{1}}$ and $\mathbf{L}^{MS\textsc{-}SSIM}$ are respectively represented by:

\begin{equation}
\mathbf{L}^{l_{1}}(x_{f}, x_{r})=\left \| x_{r}-x_{f} \right \|_{1}^{1} \\
\end{equation}

\begin{equation}
\mathbf{L}^{MS\textsc{-}SSIM}(x_{f}, x_{r})=1-MS\textsc{-}SSIM(x_{f}, x_{r})
\end{equation}
where $MS\textsc{-}SSIM$ is the multi-scale structural similarity index of the image \cite{ZWang2003}. To circumvent the loss of the image textures and details, $\mathbf{L}^{per}$ is used to generate the image having high perceptual similarity with realistic images. Moreover, $\mathbf{L}_{G}^{adv}$ stimulates our network to focus on the solutions that lie on the manifold of natural images. The computing of $\mathbf{L}^{per}$ is presented by:

\begin{equation}
\mathbf{L}^{per}(x_{f}, x_{r})=\frac{1}{K}\sum_{i=1}^{W_{u}}\sum_{j=1}^{H_{u}}\sum_{k=1}^{C_{u}}(\mathbf{\Psi} (x_{f})_{ijk}-\mathbf{\Psi} (x_{r})_{ijk})^{2}
\end{equation}
where $K=W_{u}H_{u}C_{u}$. Additionally, $W_{u}$, $H_{u}$ and $C_{u}$ denote the dimensions of the feature map $\Psi$ extracted from the DenseNet network.

\subsection{Training of A Discriminator}

$\mathbf{D}$ aims to focus on local raindrops regions, where it is likely to generate fake details and features. Similar to the generator, the discriminator also utilizes the attention maps, $\mathbf{M}_{R}$ and $\mathbf{M}_{E}$, for maximally discriminating between two probability distributions of fake and real images. $\mathbf{D}$ consists of AAM and AuAM that are helpful to update prior information from the attention maps, as seen in Figure \ref{fig_resnet}. Moreover, for stabilizing the training of $\mathbf{D}$, we utilize the novel weight normalization method that is termed spectral normalization \cite{Miyato2018}. 

\subsection{Training of Estimation Networks}

To train EstNet, we use the $\mathbf{L}^{l_{1}}$ loss function to optimize the deraining performance. The loss function of EstNet is represented by:

\begin{equation}
\mathbf{L}^{l_{1}}_{est}=\left \| x_{r}-x_{dr}^{est} \right \|_{1}^{1} \\
\end{equation}
where $\mathbf{I}_{dr}^{est}$ and $\mathbf{I}_{gt}$ are the drained image and its corresponding ground-truth image. The raindrops estimation map $\mathbf{M}_{R}$ can be generated by:

\begin{equation}
\mathbf{M}_{R} = x_{ra} - x_{dr}^{est}
\end{equation}
where $x_{ra}$ is the original rain image.

\section{Model Architecture}

\subsection{Model Architecture of The Generator}

$\mathbf{G}$ is a novel deep learning-based architecture, namely, the dilated residual dense UNets (DRD-UNet). DRD-UNet combines the advantages of dilated residual networks \cite{FYu2017}, and UNets \cite{Ronneberger2015}, allowing the maximum information flow to pass through all convolutional layers in the network and probing hierarchical features for image restoration effectively. As seen in Figure \ref{fig_resnet}, DRD-UNet is a convolutional auto-encoder comprised of an encoding part and a decoding part. The encoding part includes a set of contraction blocks. We propose using a new contraction block, namely, dilated residual dense block (DRDB), which can utilize context information from the preceding contraction blocks, and fully extract sophisticated local features from all the layers within it dense local connections. Moreover, a set of dilated convolutions with different dilation factors in each DRDB can effectively aggregate multi-scale contextual information of objects overlapped by raindrops at different scales. The bottom layer of DRD-UNet includes four DRDBs with different dilation factors. We also use skip connections among these blocks to generate continuous gradient flow from the first block to the last block and diminish the negative impact of the vanishing gradients problem. DRD-UNet also includes AAM and AuAMs to update prior information. Unlike $\mathbf{G}$, EstNet is DRD-UNet without using AAM and AuAMs.  

\begin{figure*}
 \centering
  \includegraphics[width=1\linewidth]{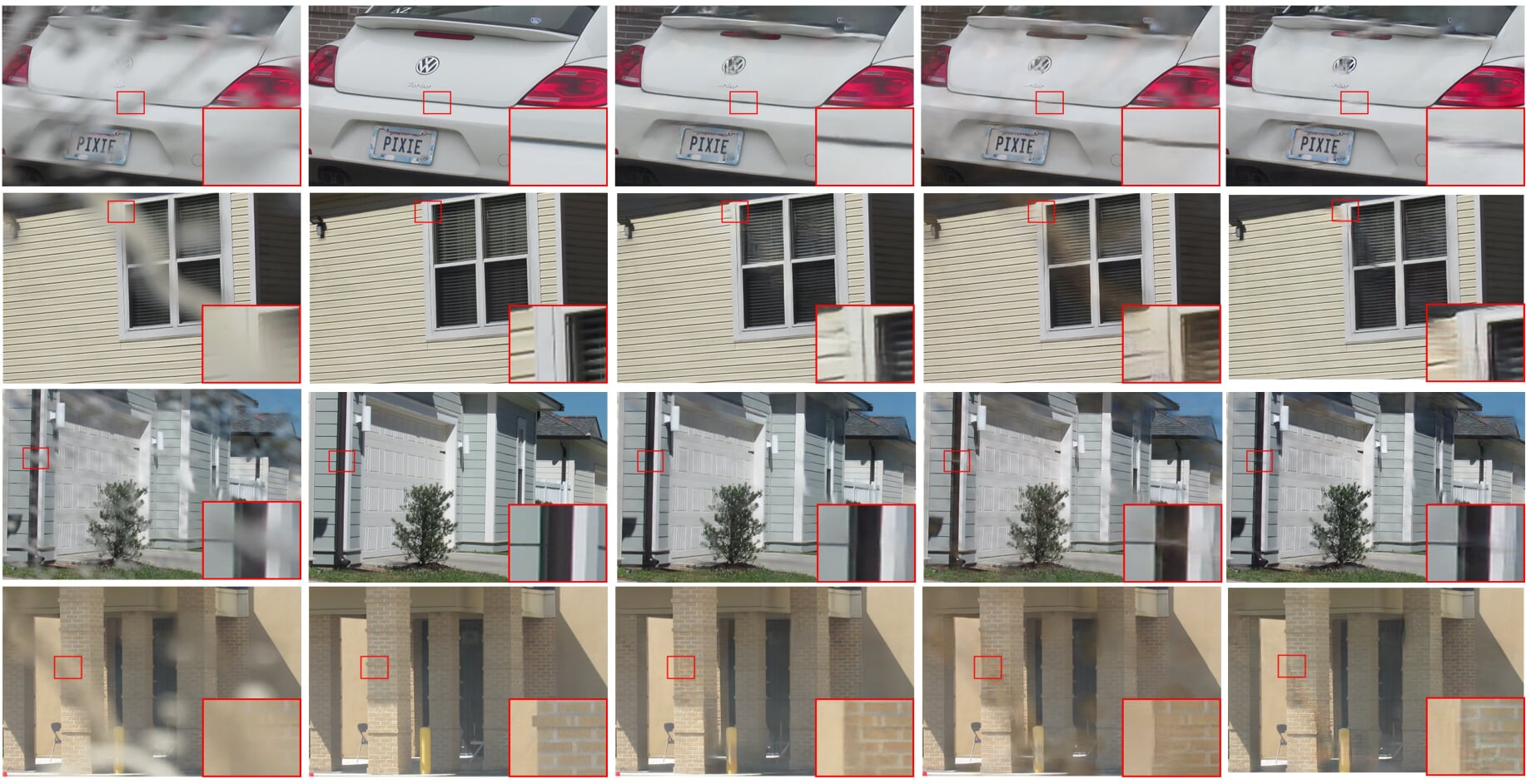} 
 \caption{Visual comparisons of different deraining methods on Deep Raindrops dataset. Column 1: Rain images. Column 2: The ground-truth images. Column 3: The results of UnfairGAN. Column 4: The results of AttenGAN. Column 5: The results of RoboCar.}
 \label{fig_derain_1}
 \vspace{-0.2in}
\end{figure*}

\begin{figure*}
 \centering
    \includegraphics[width=1\linewidth]{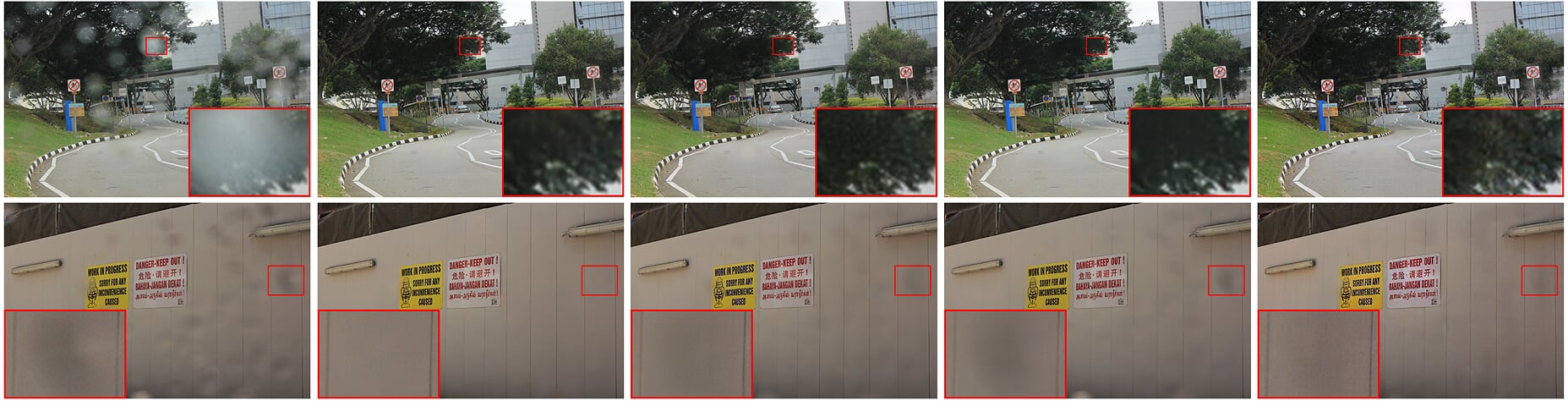} 
  \caption{Visual comparisons of different deraining methods on Raindrops dataset. Column 1: Rain images. Column 2: The results of UnfairGAN. Column 3: The results of AttenGAN. Column 4: The results of JPCA. Column 5: The results of RoboCar.}
  \label{fig_derain_2}
 \vspace{-0.2in}
\end{figure*}

\begin{figure*}
 \centering
  \includegraphics[width=1\linewidth]{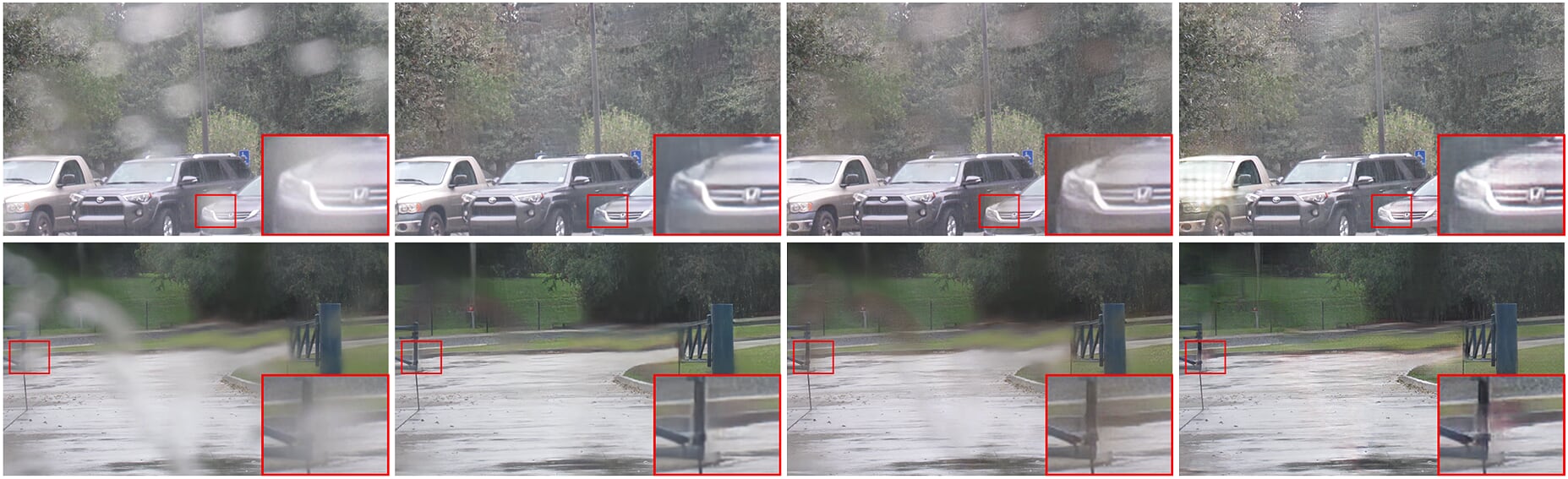} 
 \caption{Visual comparisons of different deraining methods on real-world images. Column 1: Rain images. Column 2: The results of UnfairGAN. Column 3: The results of AttenGAN. Column 4: The results of RoboCar.}
 \label{fig_derain_3}
 \vspace{-0.2in}
\end{figure*}

\section{Deep Raindrops Dataset}

Recently, there have been several datasets of raindrops \cite{RQian2018} \cite{DEigen2013} \cite{HPorav2019}, but their quality is limited and have not enough training images to sufficiently train a deep learning model. 

First, the stereo rain dataset \cite{HPorav2019} consists of 4,818 image pairs in which training, validation, and testing images were randomly selected. For this reason, the testing and training images were randomly chosen from the same video, leading to unreliable evaluations. In both testing and training images, raindrop distributions are similar, while the backgrounds are strongly correlated due to extracting from different video frames. 

Second, their raindrop dataset \cite{RQian2018} is comprised of 1,110 training and testing images. Raindrops occurring in training images from this dataset have good shapes such as circle and ellipse, while the real-world raindrop's shape varies in a large range \cite{SYou2013}. Moreover, the raindrop attached in the images from the Raindrop dataset is consistently small and thin. Thus, the features and details overlapped by the raindrop are blurred or slightly wiped out. These features are partly preserved and are not so difficult to recover from raindrops degradation. Furthermore, Figure \ref{fig_derain_2} shows the inaccuracy of testing images from the Raindrops dataset \cite{RQian2018}. The backgrounds in the raindrops image and the corresponding ground-truth image are not coincident. Moreover, mobile objects such as leaves and small trees appeared frequently in testing images.  Even though the movements are very small, they still negatively affect experimental results. These errors are the reason why this testing dataset is not reliable to test different deraining methods. 

\begin{figure*}
 \centering
    \includegraphics[width=1\linewidth]{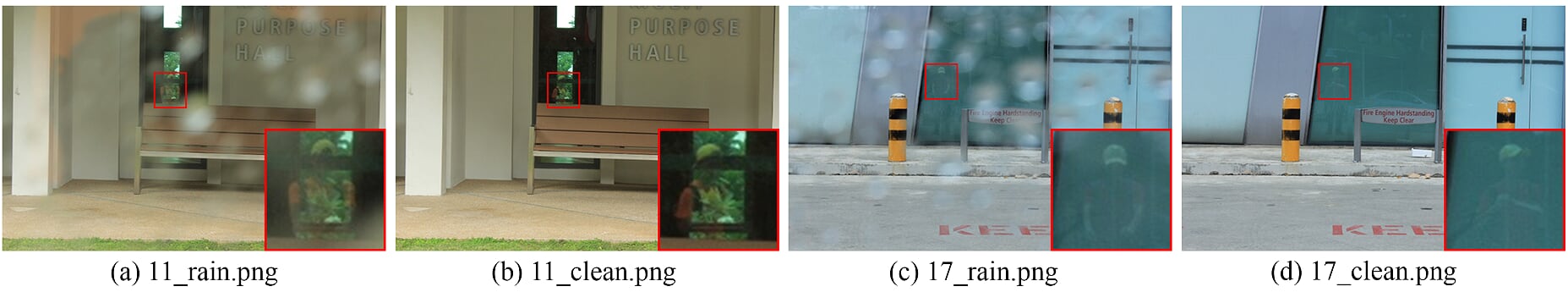} 
  \caption{Inaccuracies in the Raindrops dataset. }
  \label{fig_derain_2}
 \vspace{-0.2in}
\end{figure*}

To resolve the problem of lacking an adequate dataset for effectively training a deep learning model of removing raindrops, we built up our Deep Raindrops dataset with a large number of training pairs of raindrops and ground-truth images. To obtain rain data, we used standard cars equipped with a surveillance camera, which can quickly record a video of high-resolution 1920$\times$1080 training images. It is convenient to utilize cars for collecting data because we can use the windshield and the windscreen wiper to attach or clean water drops easily. To capture synthetic rain images, we sprayed water in the air to simulate real raindrops, which can hit the windshield at a similar speed to actual raindrops, with random directions. As a result, water drops hitting the windscreen have an equal distribution with the real-world raindrops. Indeed, raindrops will hit and spread mainly on the surface of the glass, causing unpredictable water flows as well as the complex appearance of raindrops. The realistic raindrop also causes more severe effects of refraction than that attached in the training image from the Raindrop dataset \cite{RQian2018}, leading to a challenging task of recovering the image's original details. To have high-quality training images, we had to capture images when the background objects and the car were immobile. Thus, each pair of training images has the same static background captured in various street scenes. In total, we obtained 30,000 pairs of training images and 751 pairs of testing images. In this dataset, training and testing images have a wide variety of rain rates, ranging from low rain rates to extreme rain rates. The testing dataset is divided into three categories, including Heavy-Rain, Moderate-Rain, and Light-Rain sub-datasets.

\section{Experimental Results}

\begin{table*}
\centering
\renewcommand{\arraystretch}{1.1}
\caption{The quantitative results on the Deep Raindrops dataset.}
\label{table_1}
\centering
\begin{tabular}{|p{3cm}<{\centering}|p{1cm}<{\centering}|p{1cm}<{\centering}|p{1cm}<{\centering}|p{1cm}<{\centering}|p{1cm}<{\centering}|p{1cm}<{\centering}|}
\hline
\multicolumn{1}{|c|}{Method}&\multicolumn{2}{|c|}{Heavy-Rain}&\multicolumn{2}{|c|}{Moderate-Rain}&\multicolumn{2}{|c|}{Light-Rain}
\\\hline
 Rainy & 19.55 & 0.7757 & 22.44 & 0.8418 & 25.27 & 0.8917
\\\hline 
 $\mathbf{U}$ & 25.82 & 0.8512 & 28.53 & 0.9014 & 31.33 & 0.9310
\\\hline 
$\mathbf{U\textsc{+}D}$ & 27.32 & 0.8725 & 29.59 & 0.9084 & 32.35 & 0.9399
\\\hline 
$\mathbf{U\textsc{+}D\textsc{+}G}$ & 27.37 & 0.8747 & 29.66 & 0.9109 & 32.46 & 0.9417
\\\hline 
$\mathbf{U\textsc{+}D\textsc{+}ReLU\textsc{+}G}$ & 27.62 & 0.8799 & 30.08 & 0.9172 & 32.92 & 0.9454
\\\hline 
$\mathbf{U\textsc{+}D\textsc{+}ReLU\textsc{+}UG}$ & 27.69 & 0.8801 & 30.16 & 0.9174 & 33.01 & 0.9456
\\\hline 
$\mathbf{U\textsc{+}D\textsc{+}XU\textsc{+}UG}$ & 27.87 & 0.8836 & 30.19 & 0.9179 & 33.00 & 0.9465
\\\hline 
UnfairGAN & $\mathbf{28.22}$ & $\mathbf{0.8909}$ & $\mathbf{30.49}$ & $\mathbf{0.9224}$ & $\mathbf{33.26}$ & $\mathbf{0.9491}$
\\\hline \hline
Pix2pix \cite{PIsola2017} & 23.41 & 07866 & 25.40 & 0.8303 & 27.27 & 0.8640
\\\hline
AttenGAN \cite{YQuan2019} & 24.67 & 0.7964 & 27.12 & 0.8693 & 29.61 & 0.9145
\\\hline
RoboCar \cite{HPorav2019} & 26.18 & 0.8402 & 28.57 & 0.8965 & 30.97 & 0.9208
\\\hline
\end{tabular}
\label{table_1}
\end{table*}

\subsection{Training Settings}

Based on the algorithm of Lookahead Optimizer \cite{RZhang2019}, which was used for minimizing cost functions, all our networks were trained effectively and can achieve faster convergence. Our models were trained by using the Pytorch open-source machine learning library \cite{paszke2017automatic}. Our methods were also evaluated on an Nvidia GTX 1080 Ti graphics card. We evaluated our method and the state-of-the-art methods, including JPCA \cite{YQuan2019}, AttenGAN \cite{RQian2018}, and RoboCar \cite{HPorav2019}, by using two commonly used metrics, which are PSNR, SSIM. Moreover, we also  included recent approaches which are Eigen13 \cite{DEigen2013}, Pix2pix \cite{PIsola2017}, 3RN \cite{ZHao2019}, and MixNet \cite{RLi2020}. Additionally, we did experiments on our Deep Raindrops dataset, and the Raindrops dataset Qian \cite{RQian2018} to compare the performance between our method and its competing algorithms. To demonstrate our contributions in this paper, we also compared our whole network with some baseline networks. First, $\mathbf{U}$ denotes our network without using the discriminator, dilated convolutions, and prior information. Second, $\mathbf{U\textsc{+}D}$ denotes $\mathbf{U}$ plus dilated convolutions. Third, $\mathbf{U\textsc{+}D\textsc{+}G}$ denotes $\mathbf{U\textsc{+}D}$ plus the discriminator of SRGAN. Forth, $\mathbf{U\textsc{+}D\textsc{+}ReLU\textsc{+}G}$ denotes $\mathbf{U\textsc{+}D\textsc{+}ReLU}$ plus AAM with ReLU. Fifth, $\mathbf{U\textsc{+}D\textsc{+}ReLU\textsc{+}UG}$ denotes $\mathbf{U\textsc{+}D\textsc{+}ReLU}$ plus the discriminator of UnfairGAN. Additionally, $\mathbf{U\textsc{+}D\textsc{+}XU\textsc{+}UG}$ denotes $\mathbf{U\textsc{+}D\textsc{+}ReLU\textsc{+}UG}$ in which ReLU is replaced by XUnit \cite{IKligvasser2018}. Finally, UnfairGAN is $\mathbf{U\textsc{+}D\textsc{+}ReLU\textsc{+}UG}$ in which ReLU is replaced by DAF.

\subsection{Quantitative Evaluation}

\begin{table}
\centering
\renewcommand{\arraystretch}{1.1}
\caption{The quantitative results on the Raindrop dataset.}
\label{table_Rain2}
\centering
\begin{tabular}{|p{2.5cm}<{\centering}|p{2.2cm}<{\centering}|p{2.4cm}<{\centering}|p{1.2cm}}
\hline
Methods & PSNR & SSIM 
\\\hline
Eigen13 \cite{DEigen2013} & 17.84 & 0.8299
\\\hline
MixNet \cite{RLi2020} & 30.12 & $\mathbf{0.9268}$
\\\hline
Pix2pix \cite{PIsola2017} & 30.14 & 0.6149
\\\hline
3RN \cite{ZHao2019} & 30.17 & 0.9128
\\\hline
JPCA \cite{YQuan2019} & 31.44 & 0.9263
\\\hline
AttenGAN \cite{RQian2018} & 31.52 & 0.9213
\\\hline
RoboCar \cite{HPorav2019} & 31.55 & 0.9020
\\\hline
UnfairGAN & $\mathbf{31.56}$ & 0.9215
\\\hline
\end{tabular}
\label{table_Rain2}
\end{table}

Table \ref{table_1} shows the quantitative comparisons between UnfairGAN and state-of-the-art methods, including Pix2pix \cite{PIsola2017}, AttenGAN \cite{RQian2018}, and RoboCar \cite{HPorav2019}. JPCA \cite{YQuan2019} fails to remove raindrops on this dataset because it is based on the assumption that raindrops have good shapes such as circles and ellipse. In contrast, the raindrop's shape in this dataset has a wide variety of shapes and sizes, including the form of rain flows. 3RN \cite{ZHao2019} is also unsuccessful in removing raindrops because of the missing raindrop binary mask. Since Eigen13 \cite{DEigen2013} uses a shallow convolutional neural network with three layers, it fails to train on our dataset which is for deep learning-based methods. AttenGAN performs far worse than our method in all sub-datasets. This is because AttenGAN strongly depends on attention maps, which are naturally difficult to achieve. Since attention ground-truth maps are automatically calculated, they often lead to low quality, as explained in \cite{RLi2020}. The binary form of ground-truth maps is over-simplified, resulting in the loss of vital physical properties of raindrops, as mentioned in \cite{YQuan2019}. Consequently, the derained images often generate severe artifacts, as shown in Figures \ref{fig_derain_1} and \ref{fig_derain_2}. RoboCar is slightly worse than the baseline, $\mathbf{U}$, and far worse than UnfairGAN. It is reasonable because RoboCar and $\mathbf{U}$ have a similar network architecture. Table \ref{table_1} also shows that UnfairGAN performs much better than its baseline methods. It proves that the architecture of DRD-UNet, the adversarial loss of UnfairGAN, and AAM play vital roles in developing an effective method of removing raindrops. First, Table \ref{table_1} indicates that the PSNR values by $\mathbf{U\textsc{+}D}$ are 1.5 dB, 1.06 dB, and 1.02 dB higher than those by $\mathbf{U}$ on the Heavy-Rain, Moderate-Rain, and Light-Rain sub-datasets, respectively. It demonstrates that the architecture of DRD-UNet plays a vital role in improving reconstruction performance. Second, $\mathbf{U\textsc{+}D\textsc{+}ReLU\textsc{+}UG}$ outperforms $\mathbf{U\textsc{+}D\textsc{+}ReLU\textsc{+}G}$ and improves deraining performance in both PSNR and SSIM values. This proves that $\mathbf{UnfairGAN}$ can generate images more similar to natural ground-truth images than conventional GAN. Third, Table \ref{table_1} indicates that the PSNR values by UnfairGAN are 0.75 dB, 0.83 dB, and 0.80 dB higher than those by $\mathbf{U\textsc{+}D\textsc{+}G}$, respectively. These results prove that AAM plays a crucial role in boosting restoration performance. Finally, we evaluated the performance of DAF by comparing it with other state-of-the-art activation functions, including ReLU and XUnit, as indicated in Table \ref{table_1}. Particularly, the PSNR values by UnfairGAN are 0.35 dB, 0.30 dB, and 0.26 dB higher than those by $\mathbf{U\textsc{+}D\textsc{+}XU\textsc{+}UG}$, respectively. It means that DAF significantly outperforms the state-of-the-art activation functions, XUnit, in improving image restoration accuracies. Table \ref{table_Rain2} shows the quantitative evaluation between UnfairGAN and its competing methods on the Raindrops dataset. This table indicates that UnfairGAN slightly outperforms the other methods. In this dataset, ground-truth raindrop masks are not very accurate because the binarization of the mask is not precise, and the alignment of the pair of the raindrop image and the corresponding ground-truth image is not accurate, as explained in \cite{YQuan2019}. Moreover, the number of training images is limited and is not suitable for training deep learning-based methods. For both reasons, the evaluations of competing methods on this dataset are not clear enough.

\subsection{Qualitative Evaluation}

Unlike other datasets, raindrops and rain flows are both appeared in rainy images from Deep Raindrops dataset. Figure \ref{fig_derain_1} shows that AttenGAN tends to produce artifacts, while RoboCar fails to preserve some textures and details. Figure \ref{fig_derain_2} shows the visual comparisons of the competing methods on the Raindrops dataset. This figure demonstrates that UnfairGAN is more effective than JPCA, RoboCar, and AttenGAN in removing raindrops, artifacts, and preserving details. To test our results, we invited 30 students to evaluate the competing methods. All of them chose our methods as the best one. We also demonstrated our method and its competitors on real rainy photos, collected in the same way as those on the Deep Raindrops dataset. In a real rainy photo, all the objects are wet, and the background contains rain streaks and raindrops at the same time. However, raindrops distribution in the real rainy image is highly similar to raindrops distribution in our training images. As a result, our method performs well in real-world rain images in the conditions of heavy rain, moderate rain, or light rain, shown in Figure \ref{fig_derain_3}. Moreover, the results in real photos are similar to the results of previous experiments.

Figure \ref{fig_derain_1} also shows the quantitative comparisons between UnfairGAN and its baseline methods. $\mathbf{U\textsc{+}D}$ performs significantly better than $\mathbf{U}$ due to the combination between dilated residual blocks and UNets. However, both $\mathbf{U}$ and $\mathbf{U\textsc{+}D}$ tend to smooth out the image details and textures and generate blurry local regions. In contrast, the methods using the adversarial loss of UnfairGAN preserve essential details and features more effective than $\mathbf{U}$ and $\mathbf{U\textsc{+}D}$. Moreover, by updating useful prior information of edge and raindrops, $\mathbf{U\textsc{+}D\textsc{+}ReLU\textsc{+}G}$, $\mathbf{U\textsc{+}D\textsc{+}ReLU\textsc{+}UG}$, $\mathbf{U\textsc{+}D\textsc{+}XU\textsc{+}UG}$ are even better than  $\mathbf{U\textsc{+}D\textsc{+}G}$ in erasing different kinds of rain and recovering texture details. Finally, Figure \ref{fig_derain_1} also shows UnfairGAN is the best method among them, owing to the effectiveness of DAF. To demonstrate our method, we also asked 30 students to evaluate the methods. In the final result, all of them evaluated UnfairGAN as the best method in providing high-quality images.

\begin{figure*}
 \centering
    \includegraphics[width=1\linewidth]{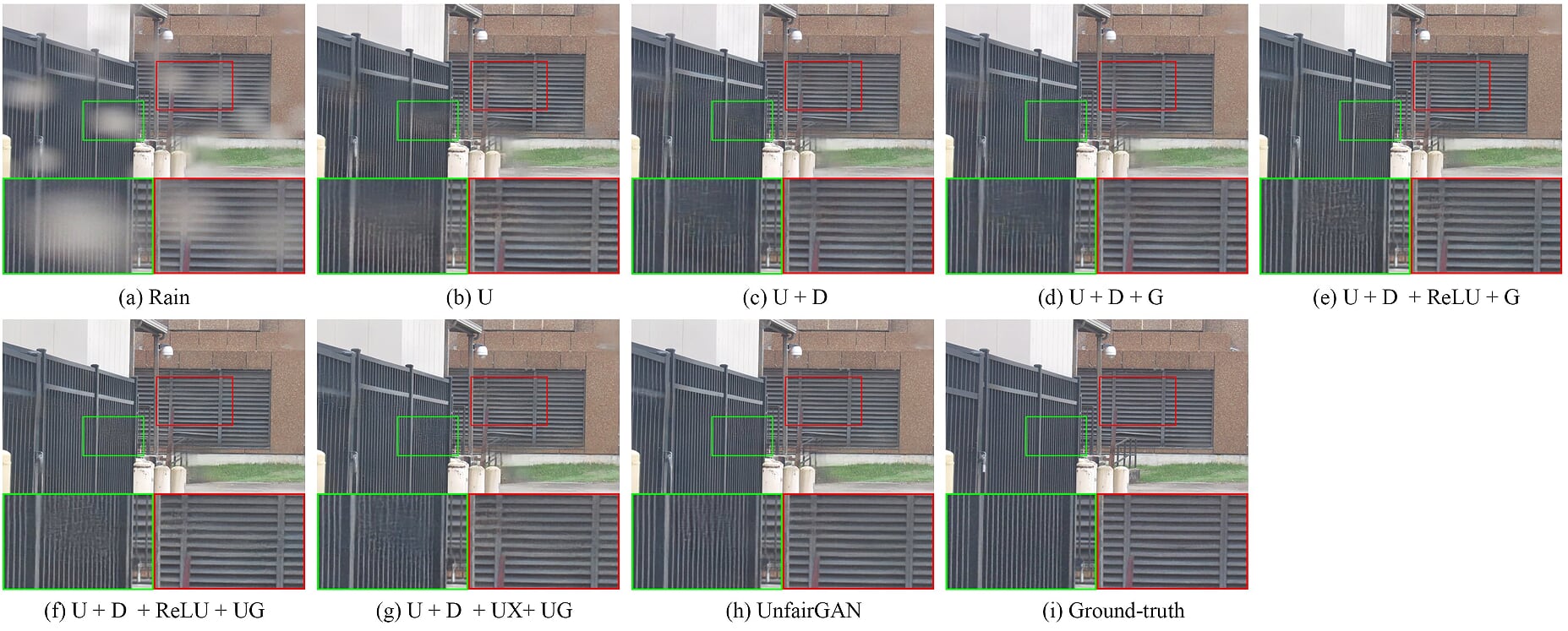} 
  \caption{Visual comparisons of different baseline methods on Deep Raindrops dataset. }
  \label{fig_derain_1}
 \vspace{-0.2in}
\end{figure*}

\section{Conclusion}

We proposed UnfairGAN to deal with the challenging problems of raindrops hitting windshields or glasses. Extensive experimental results demonstrated that UnfairGAN performed considerably better than state-of-the-art methods in handling real-world raindrops and rain flows. In this paper, we demonstrate several vital contributions to solving the recent problems of deraining, including DRD-UNet, AAM, DAF, and an improved adversarial loss.

\section*{Acknowledgements}

This research was supported by the Basic Science Research Program through the National Research Foundation of Korea funded by the Ministry of Education, Science and Technology(NRF-2015R1D1A1A01060422) and also by the MSIP (Ministry of Science, ICT and Future Planning), Korea, under the National Program for Excellence in SW supervised by the IITP (Institute for Information and Communications Technology Promotion (2015-0-00932).

\end{document}